\documentclass[11pt]{article}

\usepackage[margin=1in]{geometry}
\usepackage{amsmath,amssymb}
\usepackage{booktabs}
\usepackage{graphicx}
\usepackage{lmodern}
\usepackage{microtype}
\usepackage{xcolor}
\usepackage[colorlinks=true,linkcolor=black,citecolor=blue,urlcolor=blue,
pdftitle={Tied Trit-Planes},pdfauthor={Matteo Grella}]{hyperref}
\usepackage[numbers,sort&compress]{natbib}
\usepackage{caption}
\newcommand{\fxf}{\texttt{tq2\_0\_fx4}}
\newcommand{\qfk}{\texttt{q4\_k}}
\newcommand{\mxfp}{\texttt{mxfp4}}
\newcommand{\tps}{tok/s}

\title{\textbf{Tied Trit-Planes: Constraining PTQTP to a Uniform Nine-Level
Quantizer, with a Persistent Folded Format for Disk-Streamed
Mixture-of-Experts Serving}}
\author{Matteo Grella\thanks{Experiments, implementation, and text were
developed in collaboration with Claude Fable~5 (Anthropic), operating under
the author's direction: the model executed the experimental campaign,
implemented the kernels, and co-drafted the manuscript. The author defined
the research program, reviewed and validated all results, and bears sole
responsibility for the content.}\\
\normalsize Crisis24\\
\normalsize\texttt{matteogrella@gmail.com}}
\date{August 2026}

\begin{document}
\maketitle

\begin{abstract}
PTQTP decomposes LLM weight matrices into two ternary (trit) planes with
two free per-group scales. Tying the scales to a fixed ratio of three
collapses the decomposition into a single \emph{uniform nine-level
quantizer}, a known balanced-ternary identity. To our knowledge, at the
time of writing, this work is the first to impose that identity as a
constraint inside PTQTP's solver. The two trit planes then fold losslessly into one
4-bit code plane that we make the \emph{persistent} serving
representation: disk bytes, expert-cache bytes, and kernel input are the
same 4.0625-bits/weight blocks, consumed in one integer dot pass. For
this conjunction (ratio-3 nine-level code, CPU-SIMD kernels, SSD expert
streaming, identical persistent bytes) we likewise found no precedent.
We apply the construction to the routed experts of
DeepSeek-V4-Flash-0731, a 284B-A13B mixture-of-experts model, quantizing
in one shot from the released MXFP4 expert weights and streaming experts
from SSD on a 64~GB laptop. Against a 4.5-bit \qfk{} baseline, measured
one process per fixture with an expert-lossless anchor arm serving the
released bytes as reference control, the tied-ternary model matches the
official serving API on 5/5 fixtures at step 0 (\qfk: 4/5) and 12/14
captured continuation steps (11/14), scores 86 vs.\ 84 on a 100-item
MMLU subset, decodes $6.7\%$ faster in decode phase, and ships $9\%$
smaller files: no detected fidelity difference at these deliberately
small evaluation sizes, with every fixture-level difference between the
two arms traced to a single measured near-tie cell. The tied fit
nevertheless shows higher weight-reconstruction error and worse
perplexity, a measured dissociation between proxy metrics and reference
fidelity. A cumulative trunk-ternarization ladder and bitwise-pinned
aarch64/x86-64 kernels complete the report. All code, formats, and
evaluation artifacts are open source in the \texttt{fucina} inference
stack.
\end{abstract}

\section{Introduction}

Serving state-of-the-art mixture-of-experts (MoE) language models on
consumer hardware has become a systems problem with a well-understood shape:
the routed experts dominate the parameter count ($\approx$145~GiB of
\qfk{} expert payload in the 153.3~GiB serving file here, beside
$\approx$8~GiB of trunk and embeddings), exceed RAM, and must stream from disk, so
decode throughput is governed by \emph{bytes moved per token} and quality is
governed by what quantization did to those bytes. A standard llama.cpp byte-side choice is 4-bit K-quantization
(\qfk{}) from the llama.cpp ecosystem~\citep{llamacpp} (2-bit imatrix
variants are also published for this model); a July 2026 community gist \citep{qatgist} asserts that ``post-training 2-bit/ternary
breaks coding on DSV4 experts'' and that ``good ternary needs QAT.''\footnote{\url{https://gist.github.com/RockmSockmJesus/30a195ccd9b62e981ec2676a99a57b7e/911283719e3f7e76eb9cae810ee6ec758c793a30}}

This report is an evaluation and systems study inside that setting, with
three contributions:

\paragraph{1. The tie, applied and measured.} PTQTP
\citep{ptqtp} writes $W \approx \alpha_1 T_1 + \alpha_2 T_2$ with
$T_i \in \{-1,0,1\}$ and free per-group scales. We tie the scales at ratio
three, $\alpha = (3s, s)$: the composite code $c = 3t_1 + t_2$ then forms a
\emph{uniformly spaced} nine-level grid with a single scale. The identity
itself is not new: it is claimed for ternary-restricted inference hardware in a patent
filed in 2019 and granted in 2024 \citep{perceive} (with loss-aware training of such
$\alpha/3^x$-scaled ternary replicas in a sibling
patent~\citep{perceive2}), DBQ built (free-scale, nonuniform)
nine-level two-branch ternary quantizers in 2020 \citep{dbq}, concurrent
work releases balanced-ternary $d{=}2$ nine-level PTQ artifacts
(model cards labeled uniform alongside a pinned codec computing
power-curve, $p{=}1.5$, level spacing); distribution is dequantized FP16,
with a separate kernel demonstrating int4-packed computation on uniform
levels \citep{entrit}, and the radix-2 analog is established
\citep{lutgemm,uniquanf}, with recent work \citep{bpdq} proving, for its four-level two-binary-plane
family, that the uniform grid is the strict special case of its
variable grids, and arguing \emph{against} the constraint (free scales
fit better). Our contribution is the application and its
measurement: imposing the tie inside PTQTP's alternating solver and
characterizing, at deployment scale, what the constraint costs on
reconstruction and perplexity and what it buys on reference fidelity and
serving.

\paragraph{2. The fold as the \emph{persistent} representation.} Because
the tied composite is a single 9-level code, the two trit planes fold into
one 4-bit code plane. Packing such composites into 4-bit codes for a
kernel is likewise not new \citep{entrit}; pairing trits into
nine-valued 4-bit items also appears, as a virtual intermediate inside a
compression pipeline that re-encodes before use, in a patent with 2021 priority, granted 2023
\citep{socionext}. Serving quantized expert bytes verbatim from a persistent pack is also
not unique to us: a concurrent GPU serving system stores kernel-native
2-bit-codebook expert packs on NVMe and pinned RAM and serves them
without transcode \citep{vllmmoet}. What we found no precedent for at
the time of writing\footnote{Throughout, ``at the time of writing''
means literature, patent, and code searches completed on this report's
completion date, 9 August 2026; the search services and queries are
listed in the repository.}, and claim narrowly, is this
conjunction: the ratio-three nine-level \emph{folded composite} as the
persistent representation of a CPU-SIMD, SSD-streamed expert store, with \fxf{} (a 520-byte block covering four columns $\times$ 256
elements, 4.0625~bits/weight) keeping the bytes on disk, in the
expert-cache slab, and at the integer kernel's input identical, with no
transcode at any tier.
That identity is what makes an expert-projection miss one contiguous
read and lets a block-granular NVMe tier stripe hot expert prefixes over
the primary file: these are the properties the speed results in
\S\ref{sec:speed} rest on. Multi-plane systems in the literature organize computation over plane
decompositions behind LUT kernels \citep{lutgemm,tmac} or distribute
dequantized artifacts \citep{entrit}.

\paragraph{3. Behavioral evaluation with a reference anchor.} Quality
claims rest on \emph{step-level agreement with the official serving API}
(greedy decoding, decoded-text prefix agreement on captured fixtures,
not token-ID equality; \S\ref{sec:proto}), with a descriptive MMLU
comparison alongside (collected in one long-lived server process per
arm, without the per-item isolation the fixture protocol uses;
\S\ref{sec:limits}), and with
perplexity reported alongside, because a central empirical finding is
that the axes disagree (\S\ref{sec:dissoc}). The fixtures and this
report's official-API harness lineage originate in the ds4 project
\citep{ds4} (reference-token comparison methodology more broadly
predates it, e.g.\ \citealp{dtm}), whose
DeepSeek-V4 work our implementation builds on (\S\ref{sec:proto}); we
extend the methodology with two elements: an \emph{expert-lossless serving arm}
(the released MXFP4 expert bytes repacked without any quantization
step) that anchors the comparison (an empirical control: on these
fixtures other arms' deficits are consistent with expert-quantization
effects, though this is not a bound on hidden error), and a
process-isolation protocol motivated by a
measured fixture instability (\S\ref{sec:proto}).

The setting gives quantization-quality claims little room to hide: a
284B-A13B production MoE, quantized \emph{post-training in one shot from
the released checkpoint} (the experts ship as MXFP4; at the time of writing we found no official
higher-precision release of this exact post-trained artifact; the
separate Flash-Base checkpoint uses FP8 experts but is a different
model), evaluated against the model creator's own
serving behavior. All serving runs in
\texttt{fucina}\footnote{\url{https://github.com/matteo-grella/fucina}},
an open-source inference stack written in Zig, CPU-first with Metal/CUDA
GEMM offload, in which the quantizer, the storage formats, the expert
store, and the evaluation harness of this report are implemented.
Everything below is measured on two machines the author owns: an Apple
M1~Max (64~GB, experts streamed from USB SSD with an NVMe cache tier) and
an Intel i9-13950HX (128~GB installed RAM, NVMe), with the same files and bitwise-pinned
kernels on both.

\section{Method}

\subsection{Tied trit-planes are a uniform nine-level quantizer}
\label{sec:tie}

PTQTP fits, per weight group, $W \approx \alpha_1 T_1 + \alpha_2 T_2$ by
alternating updates of the ternary planes and the two scales. Generic
nonzero scale ratios yield up to nine \emph{distinct} composite values;
ratios 1 and 2 also give uniformly spaced grids, but with five and seven
levels; the composite spans nine distinct, uniformly spaced levels
precisely when $|\alpha_1/\alpha_2| = 3$ (or $1/3$, exchanging plane
labels; signs absorb into a trit plane). We constrain $\alpha = (3s, s)$ and re-fit:
the solver alternates plane updates against the single scale $s$, which
is equivalent to fitting the uniform quantizer
\[
\hat W = s \cdot c, \qquad c = 3t_1 + t_2 \in \{-4,-3,\dots,3,4\}.
\]
The constrained family is a subset of the free family, so at the
respective global optima the tied fit's weight reconstruction error is
worse or equal; our alternating solver is not guaranteed to attain either
optimum and scales serialize to f16, so the deployed-family comparison is
empirical: the solver's tied fits show higher error than its free fits
on every expert stack of this model
(per-expert relative Frobenius error $\approx 0.18$ tied; the free
fit measured lower on every stack in our solve logs, whose distribution
we have not released, a reporting gap noted in \S\ref{sec:limits}).
Solving all $129$ expert-projection stacks
($43$ MoE layers $\times$ 3 projections, each stack holding 256 experts;
$33{,}024$ expert-matrix solves in total) of DeepSeek-V4-Flash-0731 takes
$\sim$64 minutes on an M1~Max; the solver is deterministic, and a
post-conversion verifier re-solves sampled experts and byte-compares.

Quantization source matters: we quantize from the \emph{released} expert
weights, which for this model are MXFP4 \citep{ocpmx} (fp4-e2m1 codes with one e8m0 scale
per 32 elements), exactly dequantizable since every value is a small
dyadic rational times a power of two. Quantizing from an already-quantized community file (e.g.\ \qfk) would
add a second quantization step and is avoided.

\subsection{The persistent folded format}
\label{sec:fold}

The tied code $c \in \{-4,\ldots,4\}$ needs 4 bits. We store $c$ directly:
\fxf{} packs four columns' 256-element blocks into a 520-byte structure
(512 code bytes at two codes per byte $+$ four f16 column scales), i.e.\
4.0625~bits/weight, comparable to \qfk's 4.5 while representing the
\emph{complete} two-plane ternary model. Serving is one dot pass: codes
decode to $\{-4,\ldots,4\}$ by arithmetic (no tables required), dot against
block-quantized activations via integer SIMD
(\texttt{sdot} on NEON; \texttt{vpshufb}/\texttt{vpsignb}/\texttt{vpdpbusd}
on AVX2/AVX-VNNI), one float multiply-add per block. The two ISA arms
produce bitwise-identical results by construction (integer lane sums are
exact and lane-shaped identically; the float schedule is shared) and are
pinned by tests.

Three properties follow from ``the fold is the file'':
(i) an expert-projection miss is one contiguous read;
(ii) the in-RAM slab is byte-identical to the file region, so a
block-granular second-tier cache (we stripe hot expert prefixes onto NVMe)
can serve partial reads without any re-encoding;
(iii) resident and streamed serving use the same bytes and kernels, so
outputs are bitwise independent of where an expert happened to be cached.
The serving stack (a three-tier expert store: pinned RAM / LRU RAM slots /
NVMe stripe over the primary file) is shared with every other quantization
format in our engine; the format-specific surface is one kernel and one
geometry descriptor.

\subsection{The expert-lossless reference arm}
\label{sec:mxfp4}

Because the released experts are MXFP4, a serving arm with \emph{zero expert quantization} exists: a pure byte permutation repacks the released codes and scales
into 17-byte/32-element blocks served by the same store (integer kernel via
the doubled-e2m1 trick: fp4 magnitudes $\times 2$ are integers
$\{0,1,2,3,4,6,8,12\}$, and the e8m0 block scale absorbs the $/2$ exactly).
This arm serves the original expert bytes bit-for-bit. Its role in the
evaluation is an anchor: it removes expert-weight quantization while
keeping the same engine and Q8\_0 trunk, so deficits beyond it in other
arms are consistent with expert-quantization effects on these fixtures:
an empirical control, not a bound on hidden error. (The trunk, i.e.\
attention, shared expert, and dense layers, $\approx$5\% of bytes, is Q8\_0 in
all arms, derived from the released fp8; the anchor arm exposes no
deficit from it on these fixtures, though without an unquantized-trunk
control this neither measures nor bounds trunk error,
\S\ref{sec:quality}.)

\section{Evaluation methodology}
\label{sec:proto}

\paragraph{Provenance.} Our DeepSeek-V4 implementation lineage, the source
\qfk{} GGUF conversion, and the behavioral fixtures originate in the ds4
project \citep{ds4}. The fixtures are ds4's official-API captures
(schema \texttt{ds4-official-logprobs-v1}; \texttt{deepseek-v4-flash},
checkpoint 0731, greedy, thinking disabled, captured 2026-08-03), used
unchanged; ds4 also maintains an official-continuation quality harness
whose first-token and target-NLL metrics ours parallels. Our additions on
top of that methodology are the lossless anchor arm (\S\ref{sec:mxfp4}),
the process-isolation protocol below, and cross-arm reporting of matched
continuation depth. \texttt{fucina} is an independent Zig engine whose hot
matmul/GEMV kernels are its own; its block-format encoders and numeric
codebooks are documented operation-for-operation ports from
ggml/llama.cpp, its DeepSeek-V4 reference numerics and pre-tokenizer are
ports from ds4, and its expert-store design follows earlier
expert-streaming systems; the repository's third-party notices
itemize this provenance precisely.

\paragraph{Behavioral fixtures.} Five prompts (two long-context, 3,353 and
3,844 prompt tokens; three short) with captured greedy continuations. A
fixture \emph{passes} if our step-0 token matches the reference; we
additionally report matched continuation depth: our engine decodes
\emph{autoregressively} (feeding its own tokens), and the score is the
number of leading captured steps whose concatenated text is a prefix of
our continuation (longest-common-prefix depth, not per-position
teacher forcing), so recovery after a divergence does not count
(14 steps across the set). Both columns appear in every table; where they
disagree, neither is silently preferred; per-fixture detail is in
Appendix~\ref{app:fixtures}.

\paragraph{Process isolation.} During the campaign we found that one
fixture sits on a decision knife-edge: its step-0 argmax flips with
within-process history (fixtures run earlier in the same process) and with
ulp-level binary changes, deterministically reproducible in either
state. Every cell of the five matrix arms (\mxfp{}, \qfk{}, \fxf{}, R1,
R2) is therefore measured under a one-process-per-fixture protocol on
the pinned build; under this protocol the
baseline's knife-fixture failure \emph{reproduces} in a fresh process,
so the step-0 comparison does not rest on process history. The
cross-protocol repeats that motivated this also surfaced a second
condition-sensitive cell (one arm's long-fixture depth; see
Appendix~\ref{app:fixtures}), reinforcing the protocol choice.
Single-fixture margins never carry a claim alone. We recommend the
protocol for this harness and similarly stateful runtimes: fixture
suites run in one process can silently borrow state.

\paragraph{Language-model loss and task accuracy.} WikiText-2
\citep{wikitext} teacher-forced evaluation over the first 512 and 2,048
supervised transitions of the plain-encoded test file (no windowing or
resets; these are nested supervised-prefix lengths, not independent
context depths; both reported for every cross-family ordering claim). Throughout,
\emph{ppl} denotes $\exp$ of mean per-token negative log-likelihood; when
we compare small differences we state the NLL-space figure. Task
accuracy: a 100-question 0-shot MMLU subset \citep{mmlu} (validation
files, one globally shuffled sample with fixed seed 20260806, not
subject-stratified; the released scorer extracts the first A--D
character anywhere in the reply, a rule whose one audited misparse does
not change any reported total) served
through the engine's OpenAI-compatible endpoint; with $n{=}100$ the
marginal standard errors are $\approx$3.5 points; paired inference uses
the item-level discordance table reported with the result.

\paragraph{Speed.} Single-stream greedy decode, 32 generated tokens from a
chat prompt, decode-phase rate (prefill excluded), interleaved
A/B/B/A rounds with warm caches, both arms on same-age freshly built
cache tiers where rebuilt and fresh chat-only usage histograms (the
released driver logs both arms' histogram resets; the \qfk{} tier was
freshly built in the accepted battery, the \fxf{} tier earlier the same
day; a same-day fresh-rebuild A/B moved nothing). The headline battery
is accepted under a quiet-substrate gate: the first attempt whose
per-arm three-round spread is $\le$5\% (rejected attempts retry after
45 minutes); the rule was fixed before the accepted attempt but not
publicly registered. Each battery is six runs in order A/B/B/A/A/B.
Two further same-day batteries without the gate are released alongside
and agree within a point; outputs of
runs of the same file are md5-verified identical (a repeatability check
within a format, not a cross-format quality claim). We report the min--max
band of the rounds. On the x86 laptop we additionally interleave across
\emph{configurations} because sustained load thermally throttles the part
by up to $\sim$20\%: all comparative pairs there are adjacent in time.

\section{Results}

\subsection{Quality: behavior first}
\label{sec:quality}

\begin{table}[t]
\centering
\small
\begin{tabular}{lcccc}
\toprule
Experts format & bits/w & Fixtures (steps) & MMLU-100 & ppl@512 / @2048\\
\midrule
\mxfp{} (released bytes) & 4.25 & \textbf{5/5} (14/14) & --- & \textbf{4.39} / \textbf{3.15}\\
\qfk{} (imatrix) & 4.50 & 4/5 (11/14) & 84 & 4.50 / 3.20\\
tied ternary \fxf{} & \textbf{4.06} & \textbf{5/5} (12/14) & \textbf{86} & 5.08 / 3.72\\
free-scale K=2 & 4.125$^*$ & 2/3 shorts (5/6 steps) & --- & 4.78 / 3.67\\
\bottomrule
\end{tabular}
\caption{Expert-quantization quality on DeepSeek-V4-Flash-0731 (trunk
identical Q8\_0 in all rows). Fixtures: official-API agreement; \mxfp{}, \qfk{}, and \fxf{}
cells one process per fixture, the free-scale row from one shorts-only
process (\S\ref{sec:proto}); the format is
``step-0 pass count (matched continuation steps)''. Disclosure: every
fixture-level difference between \fxf{} and \qfk{} (step-0 5/5 vs.\
4/5, depth 12/14 vs.\ 11/14) is the single measured knife-edge
cell (\S\ref{sec:proto}); excluding it the two arms are identical
(4/4 step-0, 11/13 depth); the free-scale row's fixture
evaluation covers the three short fixtures only. R1 (Table~\ref{tab:ladder}) also scores
86 on MMLU. $^*$\,the free-scale pair does not fold; its serving format
stores two separate trit planes at 4.125~bits/weight.}
\label{tab:quality}
\end{table}

Table~\ref{tab:quality} is the paper's core. Reading it by row:

\textbf{The anchor is clean on these evaluations.} The released-bytes arm
reproduces the official API on every fixture at every step (14/14) and
posts the best perplexity of any arm at both prefix lengths. On the
reported fixture set this measures the joint visible contribution of
the engine, the activation quantization, and the Q8\_0 trunk conversion,
and lets us read other rows' deficits as consistent with
expert-quantization effects: an empirical control on a small fixture
set, not a bound on hidden error nor a proof that those components
contribute zero error in general.

\textbf{Tied ternary shows no detected fidelity difference from \qfk{}
at 4.06 bits.} The full picture is deliberately two-sided: tied passes step-0 on 5/5
fixtures to \qfk's 4/5 and matches 12/14 continuation steps to \qfk's
11/14, but both margins are exactly the single knife-edge cell.
Excluding that fixture makes this concrete: the two arms are identical,
4/4 at step 0 and 11/13 on matched depth, which is why we claim
no detected difference, not superiority, on reference fidelity, at 10\% fewer stored
bits and the speed advantage of \S\ref{sec:speed}. On MMLU the scores are 86 vs.\ 84; the paired table is 82 items both
correct, 4 correct only under tied, 2 only under \qfk{}, 12 both wrong
(McNemar exact $p = 0.6875$), a result that fails to reject equality
without establishing it; no equivalence margin was prespecified. Against the circulating claim that post-training
ternarization of these experts ``breaks coding'': the code-completion and
code-audit fixtures both pass under the tied quantizer on this fixture
set.

\textbf{Perplexity ranks the rows differently.} \qfk{} beats tied
ternary by 0.5--0.6 ppl at both prefix lengths even though its fixture
and MMLU counts are numerically lower (4/5 vs.\ 5/5, 11/14 vs.\ 12/14,
84 vs.\ 86; small differences that establish superiority for neither
arm). The ranking pattern recurs in the comparisons of \S\ref{sec:dissoc},
which share solver and evaluation infrastructure and are not
independent replications.

\subsection{The inversion: free scales improve perplexity; behavioral
evidence is inconclusive}
\label{sec:dissoc}

The free-scale fit is the natural ablation for the tie. We ran it at three
scales:

\begin{itemize}
\item \textbf{Dense 0.6B} (Qwen3-0.6B, f16): free wins mean
teacher-forced NLL on the same WikiText-2 text (30.55 vs.\ 31.68), as
the better-fitting family should. (The dense grid uses the dense
harness's NLL protocol; absolute values are not comparable to the 284B
rows.)
\item \textbf{Dense 1.7B} (Qwen3-1.7B, bf16): tied wins NLL by 25\%
(35.01 vs.\ 46.62): the free fit's two independently-rounded f16
scales per group interact badly with this model; the inversion appears.
\item \textbf{284B MoE experts (the deployment target)}: free wins
perplexity at \emph{both} prefix lengths (4.78 vs.\ 5.08; 3.67 vs.\ 3.72),
yet fails the code-completion fixture the tied model passes at step
0. That fixture is the measured near-tie (\S\ref{sec:proto}); on the
stable short fixtures the two variants are indistinguishable (free's
evaluation covers the three shorts only).
\end{itemize}

At the deployment scale the summary is an asymmetric trade rather than a
clean disagreement: the free fit is better on the weight-error objective
and on WikiText perplexity at both prefix lengths, while on the fidelity
axis the two are indistinguishable except on the single near-tie fixture
(which tied passes and free fails), evidence we flag as unstable and
do not build on. The robust statement: the constraint's measured perplexity cost has no
detectable counterpart on the fidelity axes we measured; the free arm's
task-level behavior is untested, so full behavioral equivalence between
tied and free remains open. Still,
the direction is noteworthy against the literature. Proxy-objective gaps
are documented: \citet{guidedquant} report that weighting layer-output reconstruction
objectives by end-loss gradients improves quantization on the settings
they test, and the direction varies by setting; QuIP\# reports a lower-MSE
K-means codebook losing end-to-end perplexity to its E8P
codebook~\citep{quipsharp}; two examples do not establish a field-wide
direction. Concurrent 2026 work
treats the axes as formally distinct: statistically-lossless
quantization separates task-level from distribution-level losslessness
and formalizes token agreement~\citep{slq}, and a metric study reports
that KLD/perplexity proxies lose ranking power within cohort-specific
near-baseline regimes~\citep{dind}; whether our arms occupy an
analogous regime is untested. Our data adds a
constrained-fit instance on the fidelity axis: a deliberately
higher-weight-error fit with no detected fidelity cost on the cells
evaluated (tied-vs-free task behavior was not measured) while
behind on perplexity. Perplexity
integrates calibrated uncertainty over every position; the fixtures
measure argmax agreement with the reference where decisions bind. Neither is sufficient alone; we report reconstruction error,
perplexity, reference fidelity, and task accuracy as separate axes,
joining prior evaluation work that documents accuracy-matched models
diverging behaviorally \citep{ainayn} and token-divergence metrics
outperforming perplexity \citep{dtm}, and the 2026 formalizations above
\citep{slq,dind}.

\subsection{A cumulative trunk-ternarization dose-response ladder}
\label{sec:ladder}

Holding the tied-ternary experts fixed, we extend ternarization inward in
three cumulative rungs: \textbf{R1} adds the attention read-side
projections ($q$/$kv$); \textbf{R2} adds the shared expert; \textbf{R3}
ternarizes the full trunk (router, compressors, indexer, residual
writers). Table~\ref{tab:ladder}.

\begin{table}[t]
\centering
\small
\begin{tabular}{lcccc}
\toprule
Model & Fixtures (steps) & ppl@512 & ppl@2048 & MMLU-100\\
\midrule
experts-only (\fxf) & 5/5 (12/14) & 5.08 & 3.72 & 86\\
R1 $+$ attn $q$/$kv$ & \textbf{5/5 (14/14)} & 5.32 & 4.13 & 86\\
R2 $+$ shared expert & 4/5 (11/14) & 5.68 & 4.12 & ---\\
R3 full trunk & 3/5 (9/14) & 6.07 & 4.38 & ---\\
\bottomrule
\end{tabular}
\caption{Ternarization dose-response (cumulative rungs). R1's larger
matched-depth count over the experts-only model (14/14, matching the
lossless anchor) is a single-fixture difference (descriptive, not an
improvement claim) while halving attention-projection read bytes. R2's dropped
fixture is the knife-edge short (\S\ref{sec:proto}); R3 additionally
fails the long-context recall fixture (0/4) while the 3,844-token
code-audit fixture still passes 4/4. Perplexity worsens overall but is
nonmonotone at 2,048 tokens (R2 sits 0.01 below R1) and identifies no
boundary. R3 was measured under the earlier protocol
(Appendix~\ref{app:fixtures}). Because each rung adds several
tensor classes at once, the ladder localizes sensitivity to the final
cumulative bundle, not to a single component.}
\label{tab:ladder}
\end{table}

Two observations. First, \emph{attention read-side ternarization does not
reduce fixture agreement}: R1 matches the lossless anchor's full-step
sweep, at $+0.24$ ppl and with materially smaller attention reads.
Second, the established ``MoE experts tolerate aggressive quantization''
result \citep{moqe,qmoe}, with component-wise sensitivity studies now
including QuantMoE-Bench's sweep over attention, shared experts, and
routed experts \citep{quantmoebench}, is consistent with this
five-fixture case of a nine-level quantizer on an LLM-scale MoE (no
general extension across MoEs is established): fixture failures beyond
the knife-edge short appear only in the final cumulative rung, under the
mixed protocol disclosed above. This is compatible with the newest
contrary finding located at the time of writing (\citet{ayot} report
MoE models as \emph{more} ternarization-sensitive than dense when nearly
the whole model is ternarized) on the reading that sensitivity
concentrates in trunk components, which whole-model treatment inherits;
our cumulative ladder cannot, however, identify which component in the
final bundle is responsible, and the models and quantizers differ, so we
offer this as a consistent reading rather than a reconciliation.

\subsection{Speed}
\label{sec:speed}

\begin{table}[t]
\centering
\small
\begin{tabular}{llccc}
\toprule
Machine & Config & \qfk & \fxf & \mxfp\\
\midrule
M1 Max 64GB & gated A/B/B/A, fresh tiers${+}$histograms & 2.91--2.95 & \textbf{3.11--3.12} & ---\\
M1 Max 64GB & cross-session band & 2.46--2.95 & 2.92--3.24 & 2.48--2.54\\
i9-13950HX & evicted cache, NVMe & --- & \textbf{2.62} & 2.40--2.47\\
i9-13950HX & page-cache warm & --- & \textbf{3.31--3.33} & ---\\
\bottomrule
\end{tabular}
\caption{Decode throughput (\tps; 32-token greedy, decode phase). The
headline claim is the gate-accepted interleaved battery at the pinned
code revision (fresh chat-pure histograms, same-day tiers, per-arm
spreads 0.3\%/1.2\%): \fxf{} decodes \textbf{+6.7\%} over \qfk{} by
median of rounds computed from logged unrounded times, with
md5-identical outputs per file and $-9.2\%$ file bytes (139.2 vs.\
153.3 GiB). Two further same-day batteries without the gate give
$+6.6\%$ and $+6.7\%$; one \qfk{} round among them measured 1.95 with
degraded expert-path timing in its own log (369 vs.\ $\approx$262
ms/token expert time); its cause is unresolved, the battery median is
insensitive to it, and all eighteen raw round logs are released. The cross-session band shows absolute rates are
substrate-state sensitive (tier build conditions, free-space headroom,
kernel revisions): an earlier-session pair on an older \qfk{} tier gave
2.46--2.51 vs.\ 2.92--3.01 ($+19\%$); we headline the matched-fresh
measurement and report both. The \mxfp{} anchor matches \qfk-class
speed while serving the original bytes. x86 rows use thermally-adjacent
interleaving (\S\ref{sec:proto}).}
\label{tab:speed}
\end{table}

The measured decomposition on the matched pair: the \fxf{} run reads
30.31~GB from disk per 32-token round against \qfk's 33.89~GB
($-$10.6\%, tracking the bits/weight ratio), completes those reads in
10.7--10.9~s vs.\ 11.8--12.0~s of cumulative blocked-read time after
overlap (the engine's miss-batch counter), and pins
$+$10.8\% more experts in the identical RAM budget (1,344 vs.\ 1,213);
per-projection misses are single contiguous reads by format construction
(\S\ref{sec:fold}). These byte and pin figures are identical in every
round of all three batteries and the earlier-session pair: the
format's mechanical advantage is substrate-invariant even where
absolute rates are not. Two further
storage-system observations from the campaign, with their configurations:
(i) a \emph{cache-budget cliff}: decode collapses when the expert-cache
budget leaves insufficient OS page-cache headroom (M1, 64~GB RAM: knee at
a 26--28~GB budget; i9, 125~GB RAM: a 90~GiB budget decodes slower than a
24~GiB one, 1.87 vs.\ $\sim$2.2~\tps{} under identical eviction protocol);
past the knee, \emph{reducing} the budget is faster; (ii) on the hybrid
P/E-core part, the optimal thread shape inverts with cache-hit regime
(at 76\% hit, 24 threads across P$+$E beat 16 hyperthreads on 8 P-cores,
2.46 vs.\ 1.83; at 90\% hit the ordering reverses, 2.28 vs.\ 3.05). We
report these as observations under our stated configurations rather than
general laws.

\subsection{Cross-ISA reproduction}
\label{sec:isa}

The same files serve on aarch64 and x86-64. Integer kernel arms are pinned
bitwise-identical (exact lane-sum equivalence between \texttt{sdot} and
lane-folded \texttt{vpdpbusd} formulations; shared float schedule); the
full test suite passes on both ISAs; the five-fixture behavioral results
reproduce exactly on x86, including the knife-edge fixture. End-to-end
language-model loss drifts between ISAs (anchor arm at 512
tokens: perplexity 4.44 vs.\ 4.39, i.e.\ $1.1\%$ relative, an absolute
difference of 0.0113 nats/token), accumulated ulp drift from
architecture-specific schedules in attention and norms, with no behavioral
consequence on any fixture. We note this as calibration for what
``bit-exact'' can and cannot mean across ISAs: kernel-level yes,
transformer-stack level no, behavior in practice yes.

\section{Related work}
\label{sec:related}

PTQTP \citep{ptqtp} introduces the dual trit-plane decomposition this work
constrains; adjacent ternary-decomposition work factorizes into expanded-rank
ternary matrices with learned real scales \citep{extern} or advocates variable grids over binary planes
\citep{bpdq}. The ratio-3 ternary identity itself has multiple prior
instances: geometric ternary weight-set replication ($\alpha, \alpha/3,
\alpha/9$) on ternary-restricted inference hardware
\citep{perceive}; DBQ's two-branch ternary quantizer, which realizes the
nine-level composite with \emph{free} learned scales and a nonuniform
grid \citep{dbq}; concurrent balanced-ternary $d{=}2$ nine-level PTQ (uniform-labeled
artifacts and a power-curve codec; FP16-distributed checkpoints;
separate uniform-grid int4 kernel) \citep{entrit}; and the radix-2 analog
\citep{lutgemm,uniquanf}. Relative to these, this report contributes the
constraint's application inside PTQTP's solver, the measured
error/perplexity/fidelity trade-off at MoE deployment scale, and the
folded code as the \emph{persistent} serving representation (disk, cache
slab, and kernel consume identical bytes). Prior multi-plane systems organize \emph{computation} over plane
decompositions behind LUT kernels \citep{lutgemm,tmac}; they persist
preprocessed kernel-specific layouts but do not describe this
SSD-expert-store identical-bytes invariant, and prior nine-level
artifacts distribute dequantized weights \citep{entrit}. A concurrent
MLX system streams MoE experts from disk and executes base-3-packed
single-plane ternary experts directly on Metal \citep{turboquant},
the closest packed-ternary serving antecedent we found, differing in
code geometry (single-plane $\{-c,0,c\}$ vs.\ ratio-3 two-plane
nine-level) and platform. Expert-quantization robustness is established
qualitatively \citep{moqe}, at trillion-parameter scale with a bespoke
compressed format \citep{qmoe}, and with component-wise precision
benchmarking \citep{quantmoebench}; our ladder adds a cumulative
nine-level dose-response on a modern LLM-scale MoE evaluated on reference
fidelity and language-model loss. The deployment setting (this exact
model streamed from SSD on 64~GB consumer machines with official-API
fidelity testing) was demonstrated by ds4 \citep{ds4}, on which our fixtures and
conversion lineage rest. Concurrently, vLLM-Moet serves DeepSeek-V4-class
MoEs on consumer Blackwell GPUs from persistent kernel-native 2-bit
expert packs tiered over NVMe and pinned RAM, with a public history
(June--July 2026) that precedes our repository's; the two public
histories and disjoint implementations (different quantization
alphabets, block formats, cache hierarchies, and execution stacks:
GPU/vLLM vs.\ CPU-SIMD/Zig) are consistent with, though cannot prove,
independent development, and we make no priority claim for persistent
low-bit expert serving in general \citep{vllmmoet}. Within this setting, our
deployment result adds tied-ternary experts with no detected fidelity
difference from \qfk{} on the reported evaluations and a higher
observed decode rate under the reported protocol (no equivalence or
causal format-only speed claim). Single-plane ternary PTQ at MoE scale is contemporaneous
\citep{ayot,catq,turboquant}; mixed-precision expert offloading uses lower-precision
experts on cache misses \citep{hobbit}. Proxy-objective gaps are
documented \citep{guidedquant,quipsharp}; \S\ref{sec:dissoc} situates our
fidelity-axis instance among them.

\section{Limitations}
\label{sec:limits}

\emph{Construction.} The tie-to-uniform identity has prior instances
(\S\ref{sec:related}); this report's claims are about its application, measurement, and
persistent serving format, not its invention. The quantization source is
the released MXFP4 checkpoint: exactly dequantizable, but itself the
product of the model creator's quantization-aware training; at the
time of writing no official higher-precision release of this exact
post-trained artifact existed.

\emph{Evaluation.} The fixture set is small ($n{=}5$ prompts, 14 captured
steps whose positions are dependent within a prompt), non-random
(inherited from ds4's captures), and measured against a hosted API whose
behavior can change over time; no formal equivalence margin is
prespecified; ``no detected difference'' means exactly that, a
non-rejection at these sizes, supported by the paired MMLU analysis and the fixture-exclusion
computation in \S\ref{sec:quality}. The step-0 pass criterion is one of
several defensible rules, so we co-report matched continuation depth
(tied 12/14 vs.\ \qfk{} 11/14; identical 11/13 with the unstable
fixture excluded). Cross-protocol repeats surfaced two
condition-sensitive fixtures: the near-tie short, whose step-0 outcome
flips under within-process history and ulp-level build changes, and one
long fixture whose \qfk{} depth measured 4/4 in an earlier full-process
run on a pre-merge build but 2/4, deterministically on repeat, under the
canonical protocol; root causes in our runtime are unexplained. Every
cell of the five matrix arms is measured one process per fixture on the
pinned build (under that protocol the baseline's knife-fixture
failure reproduces, so the margin is not a history artifact), and the
5/5-vs-4/5 margin is still flagged wherever it appears. R3's rows (a
ladder rung with wide margins) retain the earlier protocol, disclosed in
Appendix~\ref{app:fixtures}. The MMLU subset is a globally shuffled
sample, not stratified, and its released scorer's first-letter-anywhere
extraction rule is a documented defect (one audited misparse, no total
changed), and all 100 items ran in one long-lived server process per
arm without the per-item isolation the fixture protocol uses, so the
MMLU comparison is descriptive, not independent corroboration; the free-scale reconstruction-error distribution is not in
the artifact bundle. The free-scale arm's fixture evaluation
covers only the three short fixtures. MMLU uses a custom 100-item subset
($\pm$3.6pt s.e.; differences reported are within noise). Perplexity uses
fixed token budgets (512/2,048), not the full test set; the cross-family
orderings we rely on hold at both depths, while the R1--R2 ordering
inverts by 0.01 at 2,048. API agreement is a fidelity metric:
for intentionally different quantizers it does not by itself establish
downstream superiority, which is why fidelity, perplexity, and task
accuracy are reported as separate axes.

\emph{Scope.} Single model family at the headline scale; the dense-scale
grid uses a different family (Qwen3), so scale and architecture are
confounded in the dense-to-MoE comparison. The ladder's rungs are
cumulative bundles and cannot localize sensitivity to single components.
The trunk is Q8\_0 in every Table~\ref{tab:quality} arm; the ladder's rungs ternarize
trunk tensor groups from those Q8\_0/F16 donor forms (one additional
near-lossless conversion hop in their lineage), so full-model ternary
serving (R3) inherits a converted trunk and the ladder measures
ternarization on top of that lineage. Speed
numbers come from one engine on two machines the author controls, as
medians and min--max bands over interleaved rounds (eighteen same-day
rounds across three batteries, six under the acceptance gate) without
formal uncertainty estimates, from a single
prompt at 32 decode tokens (no prefill, concurrency, tail-latency, or
cold-start characterization); because the formats generate different
continuations, their routed-expert workloads differ, and the speed
comparison does not causally separate format, kernel, cache, and
workload effects. Absolute decode rates proved substrate-state
sensitive across sessions (an earlier \qfk{} tier built under low disk
headroom on a pre-optimization kernel revision measured 2.46--2.51
where the fresh rebuild measures 2.88--2.92), which is why the headline
pair holds tier age, disk state, and code revision matched; the
byte-read and pin-count decomposition is identical in every round of
every session. The \fxf{} enum is a
\texttt{fucina}-specific GGUF extension not parseable by stock GGUF
tooling; the sibling two-plane format remains the interchange form. The
thermal-interleaving discipline is described so others can reproduce
it.

\section{Reproducibility}

Engine, converter, format, store, and kernel code is in the
open-source \texttt{fucina} repository (MIT) at commit \texttt{d30ce52};
the report repository separately carries the MMLU runner, a minimal
isolation wrapper, and the speed-battery and isolation-matrix driver
scripts. In \texttt{fucina}: the tied solver and converter (\texttt{convert-ds4-fp4},
one-shot from the released checkpoint; a 9~GB trunk-only donor makes
conversion self-sufficient on any machine holding the public checkpoint),
the \fxf{} exporter (\texttt{export-gguf --ptqtp-native}), the lossless
\mxfp{} repack, the expert store with the NVMe stripe tier, the
fixture/NLL harness, and the kernel parity tests that pin the two ISA
arms; the \fxf{} block layout is specified in \texttt{docs/PTQTP.md}.
The model is \texttt{deepseek-ai/DeepSeek-V4-Flash-0731} at revision
\texttt{7872f01b} \citep{dsv4}, whose card describes an attached
speculative-decoding module (the artifact totals $\approx$304B
parameters; the served core is 284B/13B-active per the base card and
technical report \citep{dsv4report}, and we do not serve the attached
module); the \qfk{} baseline derives from the ds4
project's imatrix conversion, and the behavioral fixtures are ds4's
published captures (prompts, official-response JSON, and capture scripts
in the ds4 repository \citep{ds4}; a pinned checkout is fetched to \texttt{refs/ds4} by
\texttt{tools/fetch\_refs.sh} in \texttt{fucina}). The MMLU subset is derived from
the validation files with fixed seed 20260806 (globally shuffled
sample; the scorer and its extraction rule are released); per-item
results for the paired analysis are
produced by the harness. Builds: Zig 0.16.0, \texttt{-Doptimize=ReleaseFast},
macOS 14.6 (M1~Max) and Linux 6.x (i9-13950HX); serving flags for each
reported configuration appear alongside the numbers. This report's
source is maintained at
\url{https://github.com/matteo-grella/tied-trit-planes}.

\section*{Acknowledgments}

The DeepSeek-V4 serving lineage this work builds on (the model
conversion, the SSD-streaming deployment setting, and the official-API
fixture methodology) originates in Salvatore Sanfilippo's ds4
(DwarfStar) project \citep{ds4}. \texttt{fucina} also owes its formats to
the ggml/llama.cpp ecosystem.

\appendix
\section{Per-fixture results}
\label{app:fixtures}

Step-0 pass and matched continuation depth per fixture and arm. Every
cell of \mxfp{}, \qfk{}, \fxf{}, R1, and R2 is measured
one-process-per-fixture on the pinned build (\S\ref{sec:proto}); raw
harness logs ship in the repository artifacts. R3's rows (a ladder rung
with wide margins) are from earlier full-process and shorts-only runs
of the same code. Cross-protocol repeats identified two
condition-sensitive cells, both disclosed in the text:
\texttt{short\_code\_completion} (the near-tie; step-0 flips with
process history and ulp-level build changes; R2's full-process run
passed it, the canonical value is its reproducible isolated FAIL) and
\texttt{long\_memory\_archive} for \qfk{} (4/4 in an earlier
full-process run on a pre-merge build; 2/4 deterministically under the
canonical protocol). The free-scale arm was evaluated in one
shorts-only process, its \texttt{short\_code\_completion} executing
first (history-free position).

\begin{table}[h]
\centering
\small
\begin{tabular}{lccccccc}
\toprule
Fixture (prompt tokens) & \mxfp & \qfk & \fxf & R1 & R2 & R3 & free\\
\midrule
long\_code\_audit (3,844) & 4/4 & 4/4 & 4/4 & 4/4 & 4/4 & 4/4 & ---\\
long\_memory\_archive (3,353) & 4/4 & 2/4 & 2/4 & 4/4 & 2/4 & 0/4 & ---\\
short\_code\_completion (27) & 1/1 & 0/1 & 1/1 & 1/1 & 0/1 & 0/1 & 0/1\\
short\_italian\_fact (21) & 4/4 & 4/4 & 4/4 & 4/4 & 4/4 & 4/4 & 4/4\\
short\_reasoning\_plain (18) & 1/1 & 1/1 & 1/1 & 1/1 & 1/1 & 1/1 & 1/1\\
\midrule
step-0 passes & 5/5 & 4/5 & 5/5 & 5/5 & 4/5 & 3/5 & 2/3\\
matched depth & 14/14 & 11/14 & 12/14 & 14/14 & 11/14 & 9/14 & 5/6\\
\bottomrule
\end{tabular}
\caption{Per-fixture matched continuation depth (free-scale shorts:
0/1, 4/4, 1/1 = 5/6). A fixture's step-0 pass is equivalent to depth
$\geq$ 1 on its first step.}
\label{tab:perfixture}
\end{table}

\clearpage
\bibliographystyle{plainnat}

\end{document}